# Swarms on Continuous Data

**Vitorino Ramos**
CVRM-GeoSystems Centre,
Technical University of Lisbon, Portugal
vitorino.ramos@alfa.ist.utl.pt

**Ajith Abraham**
Oklahoma State University
Tulsa, OK 74106, USA
aa@cs.okstate.edu

**Abstract-** While being it extremely important, many Exploratory Data Analysis (EDA [21]) systems have the inhability to perform classification and visualization in a continuous basis or to self-organize new data-items into the older ones (evenmore into new labels if necessary), which can be crucial in KDD - Knowledge Discovery [10,1], Retrieval and Data Mining Systems [15,10] (interactive and online forms of Web Applications are just one example). This disadvantge is also present in more recent approaches using Self-Organizing Maps [4,22]. On the present work, and exploiting past sucesses in recently proposed Stigmergic Ant Systems [16,17] a robust online classifier is presented, which produces class decisions on a continuous stream data, allowing for continuous mappings. Results show that increasingly better results are achieved, as demonstraded by other authors in different areas [9,2].

## 1 Introduction

Synergy, from the Greek word *synergos*, broadly defined, refers to combined or co-operative effects produced by two or more elements (parts or individuals). The definition is often associated with the quote "the whole is greater than the sum of its parts" (Aristotle, in *Metaphysics*), even if it is more accurate to say that the functional effects produced by wholes are different from what the parts can produce alone [6]. Synergy is a ubiquitous phenomena in nature and human societies alike. One well know example is provided by the emergence of self-organization in social insects, via direct (mandibular, antennation, chemical or visual contact, etc) or indirect interactions. The latter types are more subtle and defined by *Grassé* as stigmergy [11,12] to explain task coordination and regulation in the context of nest reconstruction in *Macrotermes* termites. An example [3], could be provided by two individuals, who interact indirectly when one of them modifies the environment and the other responds to the new environment at a later time. In other words, stigmergy could be defined as a typical case of environmental synergy. *Grassé* showed that the coordination and regulation of building activities do not depend on the workers themselves but are mainly achieved by the nest structure: a stimulating configuration triggers the response of a termite worker, transforming the configuration into another configuration that may trigger in turn another (possibly different) action performed by the same termite or any other worker in the colony. Another illustration of how stimergy and self-organization can be combined into more subtle adaptive behaviors is recruitment in social insects. Self-organized trail laying by individual ants is a way of modifying the environment to communicate with nest mates that follow such trails. It appears that task performance by some workers decreases the need for more task performance: for instance, nest cleaning by some workers reduces the need for nest cleaning [3]. Therefore, nest mates communicate to other nest mates by modifying the environment (cleaning the nest), and nest mates respond to the modified environment (by not engaging in nest cleaning); that is stigmergy. Division of labor is another paradigmatic phenomena of stigmergy [13,19]. But by far more crucial to the present work and aim, is how ants form piles of items such as dead bodies (corpses), larvae, or grains of sand (fig.1). There again, stigmergy is at work: ants deposit items at initially random locations. When other ants perceive deposited items, they are stimulated to deposit items next to them, being this type of cemetery clustering organization and brood sorting a type of self-organization and adaptive behavior.

There are other types of examples (e.g. prey collectively transport), yet stimergy is also present: ants change the perceived environment of other ants (their cognitive map, according to *Chialvo* and *Millonas* [5]), and in every example, the environment serves as medium of communication [3]. What all these examples have in common is that they show how stigmergy can easily be made operational. As mentioned by *Bonabeau* et al. [3], that is a promising first step to design groups of artificial agents which solve problems: replacing coordination (and possible some hierarchy) through direct communications by indirect interactions is appealing if one wishes to design simple agents and reduce communication among agents. Finally, stigmergy is often associated with flexibility: when the environment changes because of an external perturbation, the insects respond *appropriately* to that perturbation, as if it were a modification of the environment caused by the colony's activities. In other words, the colony can collectively respond to the perturbation with individuals exhibiting the same behavior. When it comes to artificial agents, this type of flexibility is priceless: it means that the agents can respond to a perturbation without being reprogrammed to deal with that particular instability. In our context, this means that no classifier re-training is needed for any new sets of data-item types (new classes) arriving to the system, as is necessary in many classical models, or even in some recent ones. Moreover, the data-items that were used for supervised purposes in early stages in the colony evolution in his exploration of the search-space, can now, along with new items, be re-arranged in more optimal ways. Classification and/or data retrieval remains the same, but the system organizes itself in order to deal with new classes, or even new sub-classes. This task can be performed in real time, and in robust ways due to system's redundancy. Recently, several papers have highlighted the efficiency of stochastic approaches based on ant colonies for problem solving. This concerns for instance [8], combinatorial optimisation problems like the Traveling Salesman problem, the Quadratic Assignment problem, Routing problem, the Bin Packing problem, or Time Tabling problems. Numerical optimization problems have been tackled also with artificial ants, as well as Cooperative and Collective Robotics.

## 2 Stigmergy, Data, and Continuous Data

Data clustering is also one of those problems in which real ants can suggest very interesting heuristics for computer scientists. One of the first studies using the metaphor of ant colonies related to the above clustering domain is due to *Deneubourg* [7], where a population of ant-like agents randomly moving onto a 2D grid are allowed to move basic objects so as to cluster them. This method was then further generalized by *Lumer* and *Faieta* [14] (here after LF algorithm), applying it to exploratory data analysis, for the first time. In 1995, the two authors were then beyond the simple example, and applied their algorithm to interactive exploratory database analysis, where a human observer can probe the contents of each represented point (sample, image, item) and alter the characteristics of the clusters. They showed that their model provides a way of exploring complex information spaces, such as document or relational databases, because it allows information access based on exploration from various perspectives. However, this last work entitled "Exploratory Database Analysis via Self-Organization", according to [3], was never published due to commercial applications. They applied the algorithm to a database containing the "profiles" of 1650 bank customers. Attributes of the profiles included marital status, gender, residential status, age, a list of banking services used by the customer, etc. Given the variety of attributes, some of them qualitative and others quantitative, they had to define several dissimilarity measures for the different classes of attributes, and to combine them into a global dissimilarity measure (in, pp. 163, Chapter 4 [1]). More recently, *Ramos* et al. [16,17,18] presented a novel strategy (ACLUSTER) to tackle unsupervised clustering as well as data retrieval problems, avoiding not only short-term memory based strategies, as well as the use of several artificial ant types (using different speeds), present in those approaches proposed by *Lumer* and *Faieta* [14]. Following their results [16], in here we extend the ACLUSTER strategy into the possibility of obtaining continuous classification of data, forming the first data-mining framework of stigmergic self-organization of continuous and open-ended streams of data.

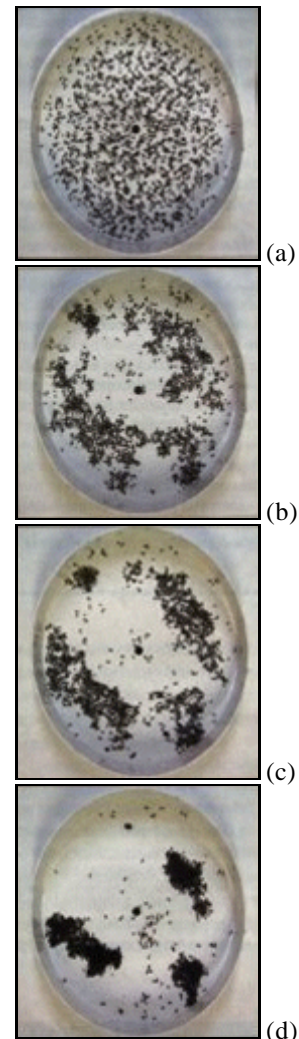
(a)
(b)
(c)
(d)

**Figure 1 -** From (a) to (d), a sequential clustering task of corpses performed by a real ant colony. 1500 corpses are randomly located in a circular arena with radius = 25 cm, where *Messor Sancta* workers are present. The figure shows the initial state (a), 2 hours (b), 6 hours (c) and 26 hours (d) after the beginning of the experiment (in [3]).

In order to confirm the validity of the model, we applied it to a collection of 244 images, each represented by a vector of 117 MM (Mathematical Morphology) extracted features [19], performing several tests of digital image retrieval. Results show that these images, on a continuous basis, are in many instances classified with rates above 90% (e.g., fig.2), which represent a successful framework for dealing with data in real time, without the overcome of re-training the classifier whenever new data (samples or even new classes) arrives to the system. These results, their comparison with overall data standard classifications, their logic in the recent Data-Mining area [15, 10], as well as the dynamic nature of the algorithm proposed are effusively discussed.

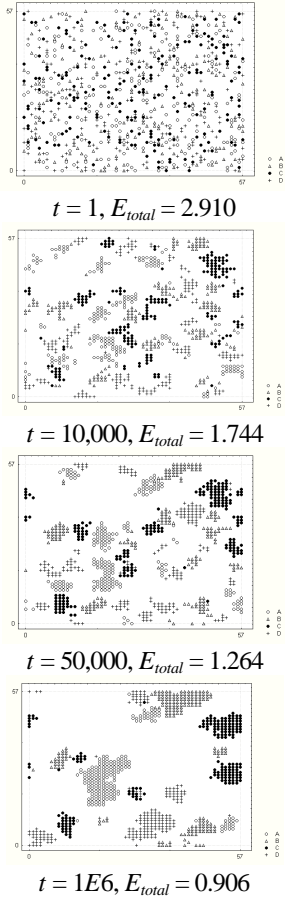

$t = 1$, $E_{total} = 2.910$

$t = 10,000$, $E_{total} = 1.744$

$t = 50,000$, $E_{total} = 1.264$

$t = 1E6$, $E_{total} = 0.906$

**Figure 2** - Some experiments with the present algorithm, conducted in artificial data (as in [14]). Spatial distribution of 800 items on a 57 x 57 non-parametric toroidal grid at several time steps. At *t*=1, four types of items are randomly allocated into the grid. As time evolves, several homogenous clusters emerge due to the ant colony action, and as expected the total entropy [16] decreases. In order to illustrate the behavior of the algorithm, items that belong to different clusters (see fig. 2), were represented by different symbols: o, Δ, • and +.

## 3 The Algorithm

The algorithm fully presented and discussed in [17] uses agents that stochastically move around the classification "habitat" following pheromone concentrations. That is, instead of trying to solve some disparities in the basic LF algorithm by adding different ant casts, short-term memories and behavioral switches, which are computationally intensive, representing simultaneously a potential and difficult complex parameter tuning, it was our intention to follow real ant-like behaviors as possible (some other features will be incorporated, as the use of different response thresholds to task-associated stimulus intensities, discussed later). In that sense, bio-inspired spatial transition probabilities are incorporated into the system, avoiding randomly moving agents, which tend the distributed algorithm to explore regions manifestly without interest (e.g., regions without any type of object clusters), being generally, this type of exploration, counterproductive and time consuming. Since this type of transition probabilities depend on the spatial distribution of pheromone across the environment, the behavior reproduced is also a stigmergic one. Moreover, the strategy not only allows to guide ants to find clusters of objects in an adaptive way (if, by any reason, one cluster disappears, pheromone tends to evaporate on that location), as the use of embodied short-term memories is avoided (since this transition probabilities tends also to increase pheromone in specific locations, where more objects are present). As we shall see, the distribution of the pheromone represents the memory of the recent history of the swarm, and in a sense it contains information which the individual ants are unable to hold or transmit. There is no direct communication between the organisms but a type of indirect communication through the pheromonal field. In fact, ants are not allowed to have any memory and the individual's spatial knowledge is restricted to local information about the whole colony pheromone density. In order to design this behavior, one simple model was adopted (*Chialvo* and *Millonas*, [5]), and extended (as in [18]) due to specific constraints of the present proposal. As described in [5], the state of an individual ant can be expressed by its position *r*, and orientation *q*. It is then sufficient to specify a transition probability from one place and orientation (*r*,*q*) to the next (*r*\*,*q*\*) an instant later. The response function can effectively be translated into a two-parameter transition rule between the cells by use of a pheromone weigthing function (Eq. 3.1):

$$W(\mathbf{s}) = \left(1 + \frac{\mathbf{s}}{1 + d\mathbf{s}}\right)^b \quad (3.1)$$

$$P_{ik} = \frac{W(\mathbf{s}_i)w(\Delta_i)}{\sum_{j/k} W(\mathbf{s}_j)w(\Delta_j)} \quad (3.2)$$

This equation measures the relative probabilities of moving to a cite *r* (in our context, to a grid location) with pheromone density *s*(*r*). The parameter *b* is associated with the osmotropotaxic sensitivity (a kind of instantaneous pheromonal gradient following), and on the other hand, 1/*d* is the sensory capacity, which describes the fact that each ant's ability to sense pheromone decreases somewhat at high concentrations. In addition to the former equation, there is a weigthing factor *w*(*Dq*), where *Dq* is the change in direction at each time step, i.e. measures the magnitude of the difference in orientation. As an additional condition, each individual leaves a constant amount *h* of pheromone at the pixel in which it is located at every time step *t*. This pheromone decays at each time step at a rate *k*. Then, the normalised transition probabilities on the lattice to go from cell $k$ to cell $i$ are given by $P_{ik}$ [5] (Eq. 3.2), where the notation $j/k$ indicates the sum over all the pixels $j$ which are in the local neighbourhood of $k$. $D_i$ measures the magnitude of the difference in orientation for the previous direction at time $t$-1.

### 3.1 Picking and Droping Data-Objects

In order to model the behavior of ants associated to different tasks, as dropping and picking up objects, we suggest the use of combinations of different response thresholds. As we have seen before, there are two major factors that should influence any local action taken by the ant-like agent: the number of objects in his neighborhood, and their similarity (including the hypothetical object carried by one ant). *Lumer* and *Faieta* [14], use an average similarity, mixing distances between objects with their number, incorporating it simultaneously into a response threshold function. Instead, in the present proposal, we suggest the use of combinations of two independent response threshold functions, each associated with a different environmental factor (or, stimuli intensity), that is, the number of objects in the area, and their similarity. Moreover, the computation of average similarities are avoided in the present algorithm, since this strategy can be somehow blind to the number of objects present in one specific neighborhood. In fact, in *Lumer* and *Faieta*'s work [14], there is an hypothetical chance of having the same average similarity value, respectively having one or, more objects present in that region. But, experimental evidences and observation in some types of ant colonies, can provide us with a different answer. After *Wilson* (*The Insect Societies*, Cambridge Press, 1971), it is knowned that minors and majors in the polymorphic species of ants *Genus Pheidole*, have different response thresholds to task-associated stimulus intensities (i.e., division of labor). Recently, and inspired by this experimental evidence, *Bonabeau* et al. [3], proposed a family of response threshold functions in order to model this behavior. According to it, every individual has a response threshold $q$ for every task. Individuals engage in task performance when the level of the task-associated stimuli $s$, exceeds their thresholds. Authors defined $s$ as the intensity of a stimulus associated with a particular task, i.e. $s$ can be a number of encounters, a chemical concentration, or any quantitative cue sensed by individuals. One family of response functions $T_q(s)$ (the probability of performing the task as a function of stimulus intensity $s$), that satisfy this requirement is given by (Eq. 3.3) [3]:

$$T_q(s) = \frac{s^n}{s^n + q^n} \quad (3.3)$$

$$c = \frac{n^2}{n^2 + 5^2} \quad (3.4)$$

$$d = \left(\frac{k_1}{k_1 + d}\right)^2 \quad (3.5)$$

$$e = \left(\frac{d}{k_2 + d}\right)^2 \quad (3.6)$$

where $n>1$ determines the steepness of the threshold (normally $n=2$, but similar results can be obtained with other values of $n>1$). Now, at $s = q$, this probability is exactly ½. Therefore, individuals with a lower value of $q$ are likely to respond to a lower level of stimulus. In order to take account on the number of objects present in one neighborhood, Eq. 3.3, was used (where, $n$ now stands for the number of objects present in one neighborhood, and $q$ = 5), defining $\chi$ (Eq. 3.4) as the response threshold associated to the number of items present in a 3 x 3 region around $r$ (one specific grid location). Now, in order to take account on the hypothetical similarity between objects, and in each ant action due to this factor, a Euclidean normalized distance $d$ is computed within all the pairs of objects present in that 3 x 3 region around $r$. Being $a$ and $b$, a pair of objects, and $f_a(i), f_b(i)$ their respective feature vectors (being each object defined by $F$ features), then $d = (1/d_{max}).[(1/F).\sum_{i=1,F}(f_a(i)-f_b(i))^2]^{\frac{1}{2}}$. Clearly, this distance $d$ reaches its maximum (=1, since $d$ is normalized by $d_{max}$) when two objects are maximally different, and $d$=0 when they are equally defined by the same $F$ features. Then, $d$ and $e$ (Eqs. 3.5, 3.6), are respectively defined as the response threshold functions associated to the similarity of objects, in case of dropping an object (Eq. 3.5), and picking it up (Eq. 3.6), at site $r$. Finally, in every action taken by an agent, and in order to deal, and represent different stimulus intensities (number of items and their similarity), present at each site in the environment visited by one ant, the strategy used a composition of the above defined response threshold functions (Eq. 3.4, 3.5 and 3.6). Several composed probabilities were analysed and used as test functions in one preliminar test [17]. The best results (fig.2 – section 2) where achieved with test function #1 below, achieving a classification rate of 94% (out of 4 different were used, as well the LM algorithm for comparison reasons – see [16,17]). However, all data in this case (244 objects corresponding to 244 images represented by 117 MM features each – fig. 2, extracted via [19]) was nourished into the system at once. What happens if we feed them continuously, starting by little groups of data ? This work tries to answer this precise question, which is not only pertinent to real world applications as can induce better results, as was demonstrated by other authors in similar areas [9,2]. For other algorithm details please consult [16,17,18].

Test Function # 1:

**Picking Probability**   **Dropping Probability**
$P_p = (1-c).e$          $P_d = c.d$

## 4 Results on Continuous Data

Instead of feeding the whole data at once to the system (as in fig. 2 [17]), the idea was to make a partition first of the data in several groups in order to test the behavior of the same algorithm on the same data (as well the final classification rate), but in contexts where the data continuously flows into the system. The 244 images of Portuguese granites [19] were then randomly separated into 6 groups. Five groups of 48 randomly samples and a final one with 4 samples only. This data was then fed into the system at several time steps. In order to compare results with [17], the total time steps used on this experiment was 1E6 as earlier. The first group of 48 samples (group A) was fed at $t=0$, the second (B) at $t=10000$, the third one (C) at $t=20000$, the fourth one (D) at $t=30000$, the fifth one (E) at $t=40000$, and finally the sixth group (F) of only four samples at $t=50000$. Now, in order to have an idea of the classification rate we measured it at several time steps and have used the same method as in [17], that is $k$-NNR - *Nearest Neighbor Rule* – i.e., a label of an unknown item is determined by the label of his first $k$ neighbors; $k=3$ was used as before. At each time step measured (see fig. 3), 20 % samples (several sub-sets were used – 10 to be precise) were randomly chosen as a test set, using $k$-NNR and the remaining samples at that time (a sort of training data) to found their labels. Using this method, figure 3 shows the average classification rate at each time step.

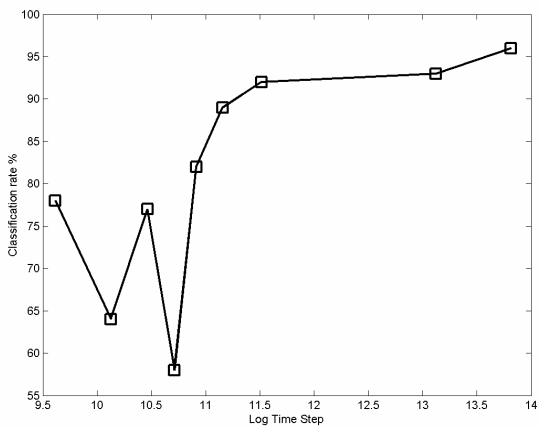

**Figure 3** – Average classification rate achieved via $k$-NNR (using 10 sub-groups of 20% samples as test and 80% as training) at several time steps (log).

## 5 Conclusions

We have presented in this paper a new ant-based algorithm for data unsupervised clustering and data exploratory analysis, while sketching a clear parallel between a mode of problem-solving in social insects and a distributed, reactive, algorithmic approach. Some of the mechanisms underlying corpse clustering, brood sorting and those that can explain the worker's behavioral flexiblity, as regulation of labor and allocation of tasks have first been introduced. As in similar past works applied to document clustering and text retrieval [16], the role of response thresholds to task-associated stimulus intensities were stressed as an important part of the strategy, and incorporated into the algorithm by using compositions of different response functions. These compositions allows the strategy not only to be more accurate relatively to behaviours found in nature as avoids short-term memory based strategies, and the use of several artificial ant types (using different speeds), present in some recent approaches. Behavioral switches as used in [14], were also avoided, in order to maintain simplicity and to avoid complex parameter settings to be performed by the domain expert. At the level of agent moves in the grid, a truly stigmergic model was introduced in order to deal with clusters of objects, avoiding randomly moves which can be counterproductive in the distributed search performed by the swarm, and adopted by all past models. In fact, the present algorithm, along with [16], was the first to introduce pheromone traces on agents to deter random explorations and encourage objects cluster formation. Moreover, we tackle the problem of having continuous data arriving to the system. In many classifier systems these represents a huge problem, since the new data not only can have different multi-dimensional distributions as can incorporate new classes, leading to dreadful results without having a new proper training. The method here described does not only avoids any type of re-training as can adapt very quickly to any new data, even if represents new classes. In fact, having continuously data can enhance the results by 2%, comparing to earlier results [17] where the final classification rate was 94%. This is achieved in part by the fact that the system can rearrange items more precisely at the first time steps, where there are few objects to handle. When the others arrived to the system, data is more easily handled since robust clusters were already found. Finally, and as verified by these tests, the present methodology indicates that a robust nonstop classifier could be achieved, which produces class decisions on a continuous stream data, allowing for continuous mappings. As we know, many categorization systems have the inhability to perform classification and visualization in a continuous basis or to self-organize new data-items into the older ones (evenmore into new labels if necessary), unless a new training happens. This disadvantge is also present in more recent approaches using Self-Organizing Maps, as in *Kohonen* maps. While a benchmark comparison of the above cited methods should be interesting to explore, the ability of this algorithm to perform continuous mappings and the incapacity of the latter to conceive it, tend to difficult any serious comparison. Suitable dynamic benchmarks of data, which can be continously fed into any classifier system should be created, in order to make serious comparisons.

**Bibliography**

[1] (1992), "Special Issue on Knowledge Discovery in Data and Knowledge Bases", *Int. Journal of Intelligent Systems*, Wiley, Vol. 7., nº. 7.

[2] A. Lumini, D. Maio and D. Maltoni (1997), "Continuous vs Exclusive Classification for Fingerprint Retrieval", *Pattern Recognition Letters*, Vol. 18, nº. 10, pp. 1027-1034.

[3] E. Bonabeau, M. Dorigo, G. Théraulaz (1999), *Swarm Intelligence: From Natural to Artificial Systems*, Santa Fe Institute in the Sciences of the Complexity, Oxford Univ. Press, New York.

[4] R. Brits, A.P. Engelbrecht (2001), "A Cluster Approach to Incremental Learning", in K. Marko, P. Werbos (Eds.), *Procs. of IJCNN'01 - INNS-IEEE International Joint Conf. on Neural Networks*, Washington D.C.

[5] D.R. Chialvo, M.M. Millonas (1995), "How Swarms Build Cognitive Maps", In Luc Steels (Ed.), *The Biology of Intelligent Autonomous Agents*, 144, NATO ASI Series, pp. 439-450.

[6] P. Corning (1998), "The Synergism Hypothesis: On the Concept of Synergy and it's Role in the Evolution of Complex Systems", *Journal of Social and Evolutionary Systems*, 21 (2).

[7] J.-L. Denebourg, S. Goss, N. Franks, A. Sendova-Franks, C. Detrain, L. Chretien (1991), "The Dynamic of Collective Sorting Robot-like Ants and Ant-like Robots", In J.A. Meyer, S.W. Wilson (Eds.), *Procs. of SAB´90 – 1st Conf. on Simulation of Adaptive Behavior: From Animal to Animats*, Cambridge, MA: MIT Press, pp. 356-365.

[8] M. Dorigo, M. Middendorf, T. Stüzle (Eds.) (2000), From Ant Colonies to Artificial Ants, *Procs. of the 2nd Int. Work. on Ant Algorithms*, Univ. Libre de Brussels, Belgium.

[9] K. Englehart, B. Hudgins, P.A. Parker, (2001), "A Wavelet Based Continuous Classification Scheme for Multifunction Myoelectric Control," *IEEE Trans.on Biomedical Engineering*, Vol. 48, No. 3, pp. 302-311.

[10] U.M. Fayyad, G. Piatetsky-Shapiro, P. Smyth, R. Uthsurusamy (Eds.) (1996), *Advances in Knowledge Discovery and Data Mining*, Menlo Park, CA, AAAI/MIT Press.

[11] P.-P. Grassé (1984), "Termitologia, Tome II", *Fondation des Sociétés. Construction.* Paris, Masson.

[12] P.-P. Grassé (1959), "La Reconstruction du Nid et les Coordinations Inter-Individuelles chez *Bellicositermes Natalensis et Cubitermes sp.* La Théorie de la Stimergie: Essai d'interprétation du Comportement des Termites Constructeurs", *Insect Societies*, 6, pp. 41-80.

[13] R.L. Jeanne (1986), "The Evolution of the Organization of Work in Social Insects", *Monit. Zool. Ital.*, 20, pp.119-133.

[14] E.D. Lumer, B. Faieta (1994), "Diversity and Adaptation in Populations of Clustering Ants". In D. Cliff, P. Husbands, J. Meyer, and S. Wilson (Eds.), *Procs. of SAB´94 – 3rd Conf. on Simulation of Adaptive Behavior: From Animal to Animats*, Cambridge, MA: The MIT Press/Bradford Books.

[15] S. Mitra, S.K. Pal, P. Mitra (2002), "Data Mining in Soft Computing Framework: A Survey", *IEEE Transactions on Neural Networks*, Vol. 13, No. 1, pp. 3-14.

[16] V. Ramos, J.J. Merelo (2002), "Self-Organized Stigmergic Document Maps: Environment as a Mechanism for Context Learning", in E. Alba, F. Herrera, J.J. Merelo et al. (Eds.), *Procs. of AEB´02 - 1st Spanish Conference on Evolutionary and Bio-Inspired Algorithms*, Mérida Univ., Spain, pp. 284-293.

[17] Vitorino Ramos, Fernando Muge, Pedro Pina, "Self-Organized Data and Image Retrieval as a Consequence of Inter-Dynamic Synergistic Relationships in Artificial Ant Colonies", in Javier Ruiz-del-Solar, Ajith Abraham and Mario Köppen (Eds.), <u>Frontiers in Artificial Intelligence and Applications</u>, Soft Computing Systems - Design, Management and Applications, <u>2nd Int. Conf. on Hybrid Intelligent Systems</u>, <u>IOS Press</u>, Vol. 87, ISBN 1 5860 32976, pp. 500-509, Santiago, Chile, Dec. 2002.

[18] V. Ramos, F. Almeida (2000), "Artificial Ant Colonies in Digital Image Habitats - A Mass Behaviour Effect Study on Pattern Recognition". In [8], pp. 113-116.

[19] V. Ramos, P. Pina, F. Muge (1999), "From Feature Extraction to Classification: A Multidisciplinary Approach applied to Portuguese Granites", in B.K. Ersboll, P. Johansen (Eds.), *Procs. of SCIA´99 - 11th Scandinavian Conf. on Image Analysis*, Kangerlussuaq, Greenland, Vol.2, Pattern Recognition Society of Denmark, pp. 817-824.

[20] G.E. Robinson (1992), "Regulation of Vision of Labor in Insect Societies", *Annu. Rev. Entomol.* 37, pp. 637-665.

[21] John Tukey (1977), *Exploratory Data Analysis*, Addison-Welsey.

[22] H.P. Siemon, A. Ultsch (1990), "Kohonen Networks on Transputers: Implementation and Animation", in *Procs. Int Neural Network Conf.*, Paris, pp. 643-646.